\documentclass[conference]{IEEEtran}
\IEEEoverridecommandlockouts
\usepackage{cite}
\usepackage{amsmath,amssymb,amsfonts}
\usepackage{algorithmic}
\usepackage{graphicx,multicol}
\usepackage{textcomp}
\usepackage{xcolor}
\usepackage{multirow} 
\usepackage{amsmath} 
\usepackage{subfigure} 

\usepackage{algorithm} 
\newcommand{\cmt}[1]{{\color{blue}  \hfill \#  #1}}

\def\BibTeX{{\rm B\kern-.05em{\sc i\kern-.025em b}\kern-.08em
    T\kern-.1667em\lower.7ex\hbox{E}\kern-.125emX}}
\begin{document}

\title{\textsc{XSub}: Explanation-Driven Adversarial Attack against Blackbox Classifiers via Feature Substitution}


\author{\IEEEauthorblockN{Kiana Vu}
\IEEEauthorblockA{\textit{Department of Cybersecurity} \\
\textit{University at Albany, SUNY}\\
Albany, NY, USA \\
kvu@albany.edu}
\and
\IEEEauthorblockN{Phung Lai\IEEEauthorrefmark{1}}
\IEEEauthorblockA{\textit{Department of Cybersecurity} \\
\textit{University at Albany, SUNY}\\
Albany, NY, USA \\
lai@albany.edu}
\and
\IEEEauthorblockN{Truc Nguyen}
\IEEEauthorblockA{\textit{Computational Science Center} \\
\textit{National Renewable Energy Laboratory}\\
Golden, CO, USA \\
Truc.Nguyen@nrel.gov}

\thanks{\IEEEauthorrefmark{1} Corresponding author}
}
\maketitle

\begin{abstract}


Despite its significant benefits in enhancing the transparency and trustworthiness of artificial intelligence (AI) systems, explainable AI (XAI) has yet to reach its full potential in real-world applications. One key challenge is that XAI can unintentionally provide adversaries with insights into black-box models, inevitably increasing their vulnerability to various attacks. In this paper, we develop a novel explanation-driven adversarial attack against black-box classifiers based on feature substitution, called \textsc{XSub}. The key idea of \textsc{XSub} is to strategically replace important features (identified via XAI) in the original sample  with corresponding important features from a ``golden sample" of a different label, thereby increasing the likelihood of the model misclassifying the perturbed sample. The degree of feature substitution is adjustable, allowing us to control how much of the original sample's information is replaced. This flexibility effectively balances a trade-off between the attack’s effectiveness and its stealthiness. \textsc{XSub} is also highly cost-effective in that the number of required queries to the prediction model and the explanation model in conducting the attack is in $O(1)$. In addition, \textsc{XSub} can be easily extended to launch backdoor attacks in case the attacker has access to the model's training data. Our evaluation demonstrates that \textsc{XSub} is not only effective and stealthy but also cost-effective, enabling its application across a wide range of AI models.

\end{abstract}



\section{Introduction}

As Artificial Intelligence (AI)/Machine Learning (ML) has increasingly become an auspicious technology in tackling various problems in big data \cite{papadopoulos2023fndaas,vinay2023robust,liu2023cross,murad2023circular,adesokan2023neuemot}, its trustworthiness has been placed under scrutiny. Previous studies have shown that ML classification models are particularly vulnerable to adversarial attacks in which, given a sample that is correctly classified by a trained model, an adversary can add small - often imperceptible - perturbations to the sample so as to arbitrarily alter the model's output \cite{goodfellow2014explaining,madry2017towards,carlini2017towards,chen2020hopskipjumpattack,dong2018boosting,szegedy2013intriguing,ding2023black}. These perturbed samples are commonly referred to as ``adversarial examples".
In fact, it is almost always possible to construct adversarial examples given any trained models \cite{szegedy2013intriguing}, necessitating rigorous research efforts to proactively identify potential attack vectors before deployment. These adversarial attacks are often categorized as white-box or black-box. White-box attacks assume that the adversary has complete knowledge of the target model, while black-box attacks only have query access to the model. Although various black-box attacks have been proposed, most of them either rely on the transferability of white-box adversarial examples to black-box models \cite{dong2018boosting,carlini2017towards} or require many queries to the target models \cite{chen2020hopskipjumpattack,amich2022eg,cheng2018query,ilyas2018black}. This number of queries directly is an important metric to measure the cost and the stealthiness of the attacks, as an AI system may charge a fee based on the number of queries and may also raise suspicion if it receives too many queries. 

Another line of research in trustworthy AI is the field of explainable AI (XAI) which aims to address the lack of transparency in the decision-making process of ML models. Through featuring various model-agnostic explainers \cite{ribeiro2016should,lundberg2017unified,sundararajan2017axiomatic, selvaraju2017grad,ying2019gnnexplainer,shrikumar2017learning,vu2021c,vu2022neucept}, XAI has emerged as a promising pathway to adding interpretable explanations on top of the existing black-box models, helping to create more effective and human-understandable AI systems. Particularly, given an input sample and a model, a feature-based explainer would indicate the importance of each feature to the model's decision. In fact, several systems have adopted the practice of releasing an explanation together with a model's output to promote trust and transparency. 

However, previous research has shown that XAI could be a potential double-edged sword: these explanations inadvertently reveal additional information about black-box models to adversaries. These additional information could then be exploited by attackers, thereby making the models more vulnerable \cite{nguyen2023xrand,shokri2021privacy,milli2019model,miura2021megex,zhao2021exploiting,severi2021explanation}. This presents an inherent trade-off between improving transparency and keeping models secure. 

Leveraging this trade-off of XAI, we propose a new explanation-driven adversarial attack, \textsc{XSub}, against black-box classifiers.  With only access to the target model's outputs and their corresponding explanations, we demonstrate that an adversary can effectively craft adversarial examples with minimal perturbations and high success rates. Note that our attack strategy does not rely on any transferable white-box adversarial examples. Additionally, \textsc{XSub} maintains a \textit{constant query complexity}, i.e., given a data sample, the number of queries to the target model needed to find an adversarial perturbation is in $O(1)$. This gives our attack a critical advantage over other black-box adversarial attacks in terms of practicality, efficiency, and stealthiness.

The main concept behind \textsc{XSub} is to target the most important features that the model is focusing on. Specifically, given a data sample that is correctly classified by the model and for which the adversary wants to find an adversarial perturbation, the model explanation would indicate features that push the model toward making the correct classification output. From that information, the adversary strategically perturbs those features by substituting them with important features of other classes. As a result, the perturbed sample contains features that would push the model to misclassify it. Interestingly, our experimental results on image classifiers show that, in certain scenarios, our attack succeeds in finding adversarial examples with only a one-pixel substitution.

Furthermore, we take an extra step to show that \textsc{XSub} can be easily extended to a backdoor poisoning attack. Assuming that the adversary has (limited) access to the training data (e.g., via crowdsourcing), \textsc{XSub} can be used to manipulate the model's decision boundary, thereby creating a backdoor trigger that would make the model misclassify samples containing it. From these attack strategies, our research further reinforces the idea that although XAI can improve trust and transparency, it would also exact harm on the model by revealing information that can be exploited by attackers.

\noindent\textbf{Contributions.} Our key contributions in this manuscript are summarized as follows.
\begin{itemize}
    \item We propose \textsc{XSub}, an explanation-driven adversarial attack against black-box classifiers with a constant query complexity. This is achieved via a novel concept of feature substitution via model explanations.
    \item We extend \textsc{XSub} to backdoor poisoning attacks in case the adversary has some access to the training process, resulting in embedding a backdoor trigger into the model.
    \item We conduct extensive experiments to demonstrate the efficacy, efficiency, and stealthiness of our proposed attack.
\end{itemize}

\noindent\textbf{Organization.} The rest of this manuscript is structured as follows. Section II discusses related studies on explanation-driven attacks and defenses. Section III establishes the technical background and threat model. Section IV demonstrates our proposed \textsc{XSub} attack. 
Section V presents a detailed experimental evaluation and discussion of the results, while Section VI provides the final concluding remarks.


\section{Related work}

Despite their benefits for ML model transparency and trustworthiness, explanations pose security risks by enabling adversaries to uncover vulnerabilities in black-box models.
Recent works have highlighted such risks via explanation-driven attacks \cite{shokri2021privacy,milli2019model,miura2021megex,zhao2021exploiting,severi2021explanation,amich2022eg,chanda2023explainability,guan2023xgbd}.
In \cite{severi2021explanation}, SHAP \cite{lundberg2020local,lundberg2017unified} can be used to extract  important features that a malware classifier focuses on by aggregating  explanations from multiple samples. These features are then used to craft backdoor triggers, which are blended into background data to change the prediction of malware samples embedded with the same trigger during inference. However, XRand \cite{nguyen2023xrand} has mitigated  such attacks by using local differential privacy to protect these features, ensuring their indistinguishability to attackers while minimizing explanation loss to preserve the utility of the explanations. Despite the progress in this area, both the attacks and defenses have primarily focused on structured data domains like malware detection, without considering the correlation among features often present in image domains. These correlations may introduce additional vulnerabilities that adversaries could potentially exploit, such as revealing the importance of one feature through another highly correlated feature, thereby posing further risks.

Several studies have investigated explanation-driven attacks in domains with highly correlated features, such as image data and graphs \cite{amich2022eg,yan2023explanation,yan2024explanation,chanda2023explainability,guan2023xgbd}. In \cite{amich2022eg}, the authors introduce EG-Booster that utilizes feature-based explanations from image classifiers to guide the crafting of adversarial examples. They selectively add perturbations that are likely to cause model evasion while avoiding non-consequential perturbations that are unlikely to affect the model's decision. This approach leverages existing attacks to generate baseline adversarial samples and then queries the classifier multiple times to identify which consequential and non-consequential perturbations should be applied to enhance the baseline attack.
 While this approach is highly effective in white-box attacks, it performs poorly against black-box models, achieving only a $28.87\%$ evasion rate. In addition, this approach  requires multiple queries to the ML classifier to check the labels of modified samples, which limits its practicality in scenarios where querying incurs usage fees.


A recent line of work focuses on defending against adversarial and backdoor attacks by detecting and removing poisoned samples before training or testing models \cite{wang2019neural,gao2019strip,ma2022beatrix,cheng2023beagle}. In \cite{ma2022beatrix}, the authors propose the Beatrix defense mechanism, which identifies potentially poisoned  data samples by searching for anomalous patterns. Specifically, the defense examines the median absolute deviations among the inner products of feature maps between a known clean sample and a potentially poisoned one. If the deviations exceed a predefined threshold, the sample is flagged as a potential poisoned sample.

\section{Preliminaries}
In this section, we first revisit model explanations and then define the threat model for our work, essentially outlining the capabilities and goals of the adversary.

\subsection{Model Explanations}
The goal of model explanations is to enhance the transparency and reasoning of machine learning (ML) models by capturing how each feature influences the model's decisions and which class such decisions favor. Given a sample $x = \{x_j\}_{j=1}^d$ where $x_j$ represents the  $j^{th}$ feature of the sample and $d$ is the number of features, let $f$ be a model function in which $f(x)$ is the probability that $x$ belongs to a certain class $y$. 
An explanation model 
$g (x)$  is typically simpler than  $f(x)$ and easier for users to understand. For instances,  linear models and  decision trees are commonly used as  explanation models \cite{ribeiro2016should,lundberg2017unified,mahbooba2021explainable}.  


\textit{\textbf{Shapley Additive Explanations (SHAP) \cite{lundberg2020local,lundberg2017unified}.}} A SHAP explanation is based on Shapley values \cite{hart1989shapley,roth1988shapley}, which use a coalitional game theory concept to calculate the contribution of each feature to the output of the prediction model. 
 Shapley values tell us how to fairly distribute the model prediction among the features.  
SHAP explanation  is represented as a linear model, as follows: 
\begin{equation}
    g(x) = \sum_{j=1}^d e_j x_j, 
\end{equation}
where   $\{e_j\}_{j=1}^d$ are the coefficients of the explanation model $g(x)$,   measuring the impact of the feature $x_j$ on the model's decision  $f(x)$.   
Here, we consider a vector $e = [e_1, e_2,  \cdots, e_d]$ is the representation of the explanation model. 
In general, higher values of $e_{j}$ imply a higher impact of the feature $x_j$ on the model decision.

\begin{figure}[t]
 \centering
\subfigure[Car class]{\includegraphics[scale=0.425]{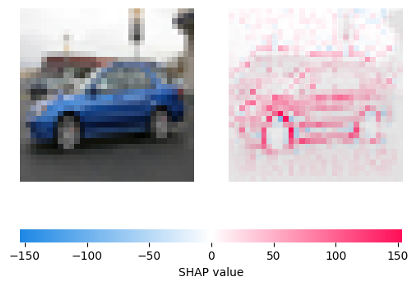}}\hfill
\subfigure[Horse class]{\includegraphics[scale=0.35]{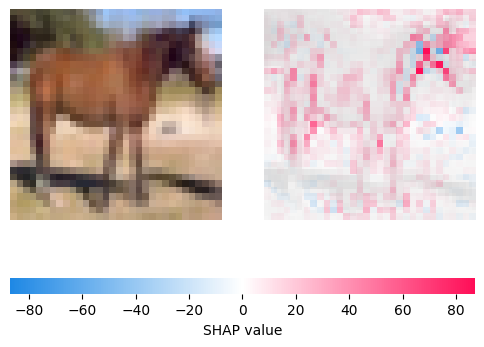}}
\subfigure[English springer class]{\includegraphics[scale=0.35]{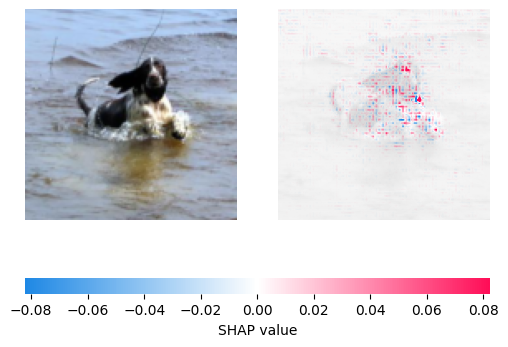}}\hfill
\subfigure[Parachute class]{\includegraphics[scale=0.35]{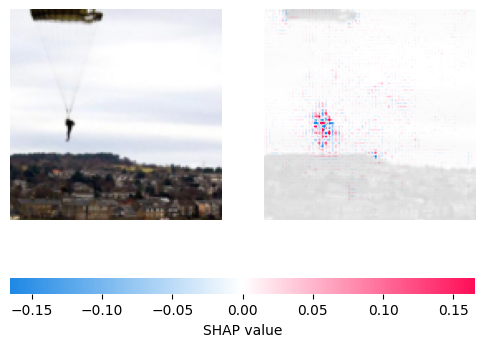}}
\caption{Examples of golden samples from the CIFAR-10 dataset \cite{krizhevsky2009learning} (upper row images) and the Imagenette dataset \cite{Howard_Imagenette_2019} (lower row images).} 
\label{fig:goldenimage}
\end{figure}
\setlength{\textfloatsep}{5pt}

\subsection{Threat Model}
In this work, we focus on an adversarial attack (i.e., inference-time) setting and further extend our attack to a backdoor attack (i.e., training-time) setting, based on 
the attacker's access to data samples. First, for an adversarial  setting, the attacker only has access to and can poison the testing  samples. Their goal is to alter the label of these samples during inference. In this scenario, the attacker can manipulate the labels of poisoned samples at inference time but cannot influence how the model is trained. Second,  for a backdoor  setting,
the attacker can inject poisoned samples into the training data, altering the training process to create a backdoored classifier that differs from a clean classifier. In this case, the attacker aims to make the model misclassify samples embedded with a trigger while ensuring that the model's responses to clean inputs remain consistent with those of the clean classifier.  
This is a practical setting for ML-as-a-Service (MLaaS), where models are trained on crowd-sourced data, thereby enabling attackers to potentially manipulate the training data.
In both settings, we consider  an \textit{untargeted} scenario, where the goal of the attacks is to misclassify the model without specifying a particular target label.




\section{\textsc{XSub}: Explanation-Driven Adversarial Attack with Feature Substitution}

\begin{figure}[t]
 \centering
\subfigure[Original image]{\includegraphics[scale=0.25]{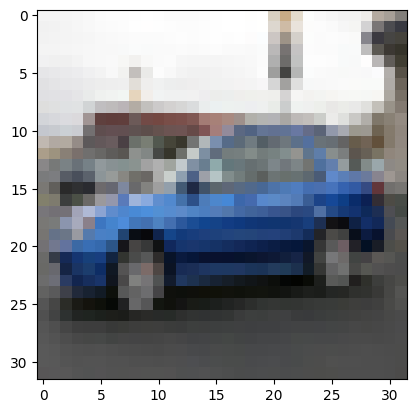}}\hfill
\subfigure[$K$=1]{\includegraphics[scale=0.25]{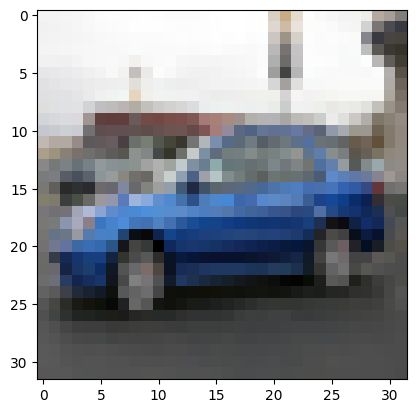}}\hfill
\subfigure[$K$=30]{\includegraphics[scale=0.25]{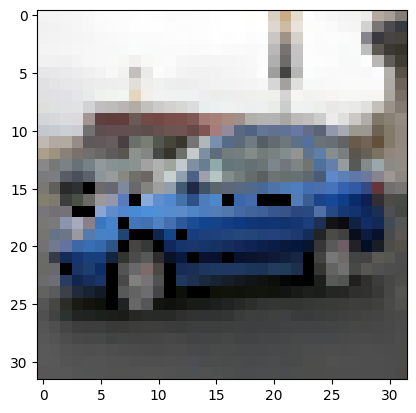}}\hfill
\caption{\textsc{XSub} with varying values of $K$ ($\alpha=\beta=100$).} 
\label{fig:varyK}
\end{figure}
\setlength{\textfloatsep}{5pt}

In this section, we introduce \textsc{XSub}, a novel explanation-driven adversarial attack tailored for black-box classifiers. The goal of \textsc{XSub} is to utilize  model explanations to guide the perturbation of data samples, while minimizing the differences between the perturbed data samples and their original. This adversarial attack  operates entirely in a black-box setting, where the attack has no access to the ML model itself. In addition, it can be easily extended to a backdoor attack if the adversary gains access to the model's training data, which is a practical settings such as ML-as-a-Service (MLaaS) \cite{minthigpen,ibm}. A shining feature of  \textsc{XSub} is its efficiency, requiring only a constant query complexity to the ML model. This feature makes it highly cost-effective in settings where access to ML models incurs usage fees such as Amazon SageMaker (AWS), Google Cloud AI Platform, Microsoft Azure Machine Learning, IBM Watson Machine Learning, etc. 


Given a sample $x \in \mathbb{R}^d$ with  label $y$ from a test set $D_{test}$, a black-box classifier $f$, and an explanation model $g$, our goal is to construct a poisoned sample $x'$ such that its label $y'$ differs from $y$ with minimal changes compared with $x$.  To achieve the goal, our idea is to identify important features based on model explanations for a given data sample, and then substituting them with those from another data sample with a different label. This raises the following question: \textit{Which data sample should be chosen for substitution, and how should the substitution be performed to optimize the trade-off between  effectiveness and stealthiness?}

To answer these questions, we introduce a concept of a \textbf{\textit{golden sample}} and a novel \textbf{\textit{explanation-driven substitution}} mechanism, as described below.

\subsection{Golden Sample Selection}
Given a set of samples $\mathbb{S}$ from a specific class $y_{\mathbb{S}} \neq y$,   the  explanation model $g$ provides an explanation vector $e_{i}=[e_{i1}, e_{i2}, \cdots, e_{id}]$ for the sample $x_i \in \mathbb{S}$.    
A golden sample $\mathcal{I}_G$ of the class $y_{\mathbb{S}}$ is defined as the sample in $\mathbb{S}$ with  the highest explanation value $e$ for that class, as follows: \vspace{-5pt}
\begin{equation}
    \mathcal{I}_G = \arg \max_{x_i \in   \mathbb{S}}e_{x_i\max} \vspace{-5pt}
\end{equation}
where $e_{x_i\max} =  \arg \max_{e_{ij}} [e_{i1}, e_{i2}, \cdots, e_{id}]$. 

Specifically, in our experiments with image classifiers, where each pixel is represented by three color channels (or two in gray-scale images), we aggregate the explanation values across channels by summing them at each pixel position. The golden images is then the one with the highest channel-aggregated explanation value for the specific class. Figure \ref{fig:goldenimage} shows examples of  golden images for several classes in the CIFAR-10 and Imagenette datasets.  

It is worth noting that the golden sample selection can be done offline, which significantly saves both time and effort. By conducting this process in advance, we can ensure that quality standards are met without disrupting workflows, allowing for smoother operations. In addition, 
this approach mitigates the need for repeated queries for a single data sample. By proactively preparing the golden sample for each class, we can optimize delays and ensures that  the golden sample is readily available when needed.


\subsection{Explanation-Driven Substitution}
After selecting the golden sample $\mathcal{I}_G$ from class $y_{\mathbb{S}} \neq y_x$, we construct a poisoned sample $x'$ from the original sample $x$ by substituting all $K$ important features contributing to the prediction of $y_x$ with the corresponding  $K$ important features of the golden sample. 
These important features are identified by the explanation vector, where higher values indicate greater importance to the model decision. The $K$ important features being substituted are referred to as the  \textit{golden positions}. 
The substitution is performed in the same order of importance within these golden positions. 
For instance, the top-$1$ important feature of $x$ is substituted by the top-$1$ important feature of the golden sample, and so on, up to the top-$K$ important feature.  It is important to note that different samples may have different golden positions, depending on the locations of their most important features. The explanation-driven substitution is formulated as follows:
\begin{equation}
\centering
    x' = x - \alpha\delta_{xK} + \beta\delta_{\mathcal{I}_GK}
    \label{eq:substitution}
\end{equation}
where  $\alpha$ and $\beta$ are positive amplification hyper-parameters. Technically, $\delta_{xK}$ and $\delta_{\mathcal{I}_GK}$ represent masks of the same size as $x$, with all elements set to zero, except for those corresponding to the golden positions of $x$ and $\mathcal{I}_G$, respectively. The non-zero values match the values of the features in their respective golden positions. For example, in our experiments with image classifiers,  setting $\alpha = \beta = 1$, we substitute pixel values at golden positions in the original image with  pixel values from the golden image.



\begin{figure}[t]
      \centering 
       \includegraphics[scale=0.026]{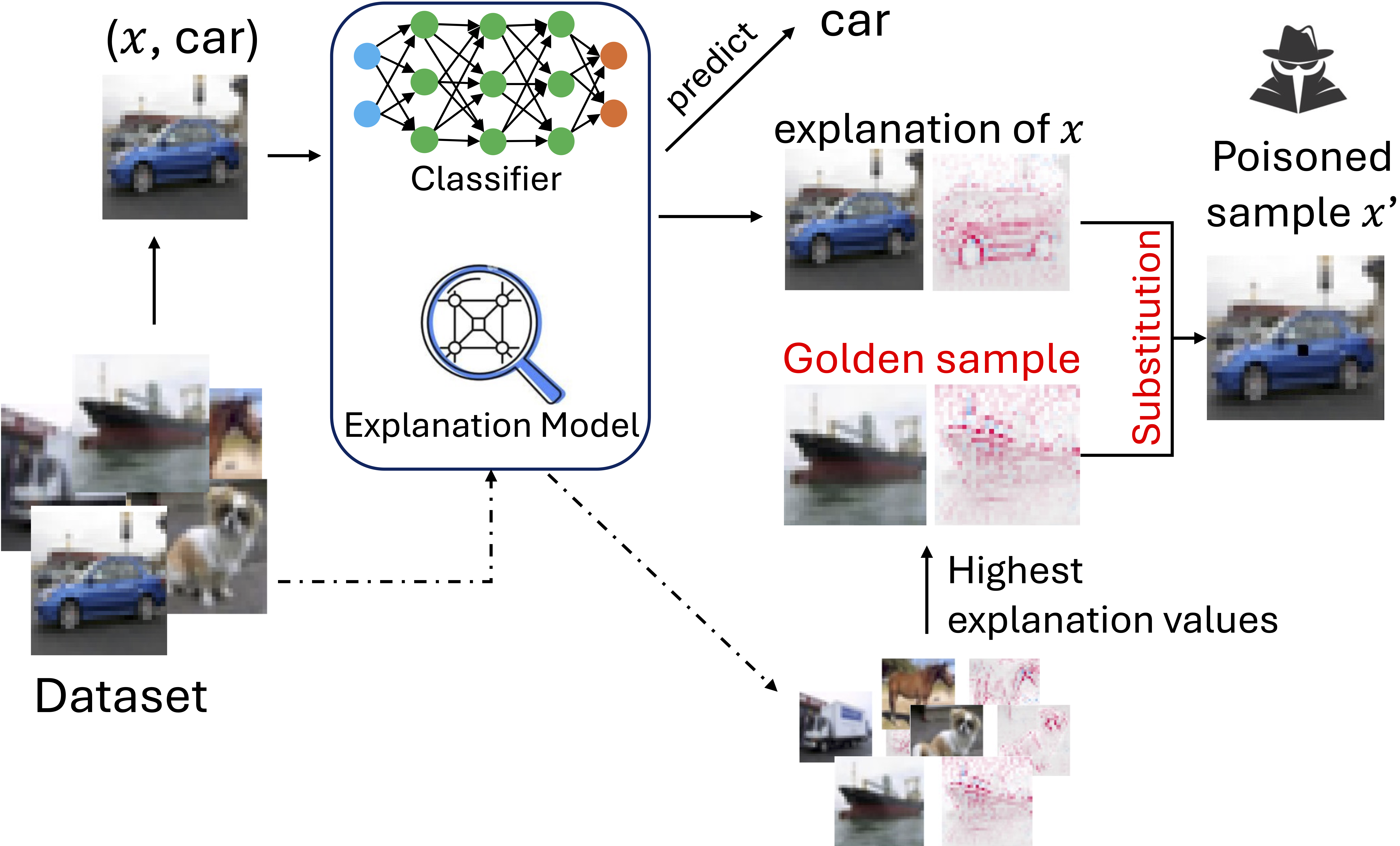} 
      \caption{The framework of our proposed attack \textsc{XSub}.}
      \label{fig:overview}
\end{figure} 

\subsection{Explanation-Driven Adversarial Attack}
As shown in Algorithm \ref{alg: LDPstealIP} and Figure \ref{fig:overview}, to perform an adversarial attack, given a data sample $(x,y_x)$ \textsc{XSub} first queries the classifier $f$ and the explanation model $g$ to obtain the prediction $\hat{y} = f(x)$ and explanation $g(x)$ (Line 4). Next, a different class 
$y_{\mathbb{S}} \neq y_x$ is randomly selected, and several test samples belonging to that class are randomly chosen to form the set $ \mathbb{S}$.  We then query to obtain the explanations $g(x_i)$ for all $x_i \in \mathbb{S}$. A golden sample $\mathcal{I}_G$ is identified based on these explanations (Lines 5 and 12-16).  With pre-defined values of  $K$, $\alpha$, and $\beta$, we substitute top-$K$ important features of the sample $x$ with the corresponding top-$K$ features of   $\mathcal{I}_G$ using Equation \ref{eq:substitution} to generate the poisoned sample $x'$  (Line 6).

\textbf{Impacts of $\alpha$, $\beta$, and $K$.}
When $\alpha$ and $\beta$ are $0$, no substitution is applied, meaning the original image remains unchanged. If $\alpha$ and  $\beta$ are set to $1$, we directly substitute the features at golden positions in the original sample with the associated features from the golden sample.   
When $\alpha$ and $\beta$ are set to higher values, this results in an amplification of the substitution effects, potentially making the adversarial perturbation more impactful in the poisoned sample.  These amplification hyper-parameters allow for fine-tuning the degree of perturbation, depending on the desired balance between stealthiness and effectiveness of the attack. 

The hyper-parameter $K$  indicates the number of features being substituted. Increasing $K$ results in more important features of the original sample being replaced,  thereby boosting the likelihood of misclassification. However, higher values of $K$ may compromise the  stealthiness of the attack, as replacing more features can make the perturbation more noticeable to human observers or defense mechanisms. This trade-off is illustrated in Figure \ref{fig:varyK}.

\textbf{Impacts of a golden sample.}
Golden positions in the golden sample with the highest important explanation values indicates the features that have the greatest influence on the model decision. By focusing on these key features, explanation-driven substitution enables the replacement of the most important features in the original sample's prediction with the corresponding important features from the golden sample. This targeted approach significantly enhances the probability of changing the label of the original sample, as it disrupts the model's confidence on the key features that were initially driving its prediction. Utilizing the golden sample, instead of a random sample, ensures the changes are subtle but impactful, effectively manipulating the model outcomes. In addition, this strategy supports to reduce the number of $K$, thereby enhancing  the stealthiness of \textsc{XSub}. 

\begin{algorithm}[t]
\small
\caption{\textsc{XSub} Algorithm}\label{alg: LDPstealIP}
\begin{algorithmic}[1]
\STATE \textbf{Inputs}: Black-box classifier $f$, training set $D$, test set $D_{test}$,  explanation model  $g$ with its explanations  $e = [e_1, e_2,  \cdots, e_d]$, loss function $\mathcal{L}$, test sample $(x, y_x)$, hyper-parameters $\alpha$, $\beta$, and $K$, percentage of data poisoning $p$
\STATE \textbf{Outputs}:  Poisoned sample $x'$, Backdoored model $f'$ (in the backdoor setting)
\STATE \textbf{Adversarial Attack}:
\begin{ALC@g}
\STATE Query the classifier $f$ and the explanation model $g$ to obtain $\hat{y} = f(x)$ and $g(x)$  
\STATE  Find a golden sample for another (random) class $y_{\mathbb{S}} \neq y_x$:  \\  \quad \quad $\mathcal{I}_G = H(y_x,f,g,D_{test})$ 
\STATE Substitute top-$K$ important features of $x$ by those of $\mathcal{I}_G$\\ \quad \quad 
$x' = x - \alpha \delta_{xK} + \beta \delta_{\mathcal{I}_GK}$
\STATE \textbf{Return:} $x'$
\end{ALC@g}
\STATE \textbf{Backdoor Attack}:
\begin{ALC@g}
\STATE Poison $p\%$ of training data to obtain $D_p$ \cmt{Similar to the Adversarial attack} 
\STATE Train $f$ with $D \cup D_p$: \\
\quad \quad $f' = \arg\min_{f} \mathcal{L}\big(f(D) + f(D_p)\big)$
\STATE \textbf{Return:} $f'$
\end{ALC@g}
\STATE \textit{\textbf{Find a golden sample $H(y_x,f,g,D_{test})$}}:
\begin{ALC@g}
\STATE Randomly select a set of data samples $\mathbb{S} = \{(x_i, y_{\mathbb{S}}) \in D_{test}\} $ where $y_{\mathbb{S}} \neq y_x$ \cmt{All samples in $\mathbb{S}$ have the same label $y_{\mathbb{S}}$}
\STATE Query $f$ and $g$ to get $\hat{y}_i  = f(x_i)$ and $g(x_i)$ for all $x_i \in \mathbb{S} $
\STATE  For all $x_i \in \mathbb{S}$:\\ \quad \quad $e_{x_i\max} =  \arg \max_{e_j, j \in [1,d]} e_{i}=[e_{i1}, e_{i2}, \cdots, e_{id}]$  
\STATE \textbf{Return:} $\mathcal{I}_G = \arg \max_{x_i \in   \mathbb{S}}e_{x_i\max}$ 
\end{ALC@g}
\end{algorithmic} 
\end{algorithm} 
\setlength{\textfloatsep}{10pt}

\subsection{Extension to Explanation-Driven Backdoor Attack}
The adversarial attack in \textsc{XSub} can be easily adapted to a backdoor attack setting by assuming that the adversary gains access to the training data of the model. This setting is practical and feasible in many real-world ML services, such as MLaaS, where  models are frequently updated using 
 crowd-sourced data.  By exploiting this access, adversaries can craft and submit  poisoned data  for model training, potentially altering the model decision boundary to respond differently when a trigger is present. In this context, we consider the substitution itself as the trigger. 
 
In \textsc{XSub}, to carry out a backdoor attack, we first poison $p$ percentage of the ML model's training data to create a poisoned training set $D_p$ (Line 9). We then submit $D_p$ to the server frequently for model training (Line 10). By doing that, we modify the decision boundary to adopt the poisoned data,  causing it to respond differently when the trigger (i.e., the substitution) appears.

\subsection{Summary of \textsc{XSub} Novelty and Benefits} 
  The novelty and benefits of \textsc{XSub} stem from its unique design, which leverages golden sample selection and  substitution to manipulate model outcomes. Here are the key advantages:
   \textbf{\textit{1)}} The use of a golden sample with the highest explanation values for substitution increases the probability of changing labels, thereby enhancing the effectiveness of the attack. In addition, this approach improves the stealthiness of the attack by minimizing the number of features $K$ that need to be replaced.
\textbf{\textit{2)}} The substitution process itself is both simple and flexible. By varying values of $\alpha$, $\beta$, and $K$, we can control how much information in the original sample will be replaced, effectively balancing the trade-off between the attack's effectiveness and its stealthiness. 
\textbf{\textit{3)}} Another standout feature of \textsc{XSub} is its efficiency. For each sample, it only requires a constant query complexity  to  the prediction model and the explanation model, making it highly cost-effective in settings where querying models incurs usage fees.
\textbf{\textit{4)}} 
\textsc{XSub} can easily be adapted into a backdoor attack by providing access to the model's training data, which is a practical scenario in many MLaaS environments. 
  As a result, \textsc{XSub} effectively balances the trade-off between attack effectiveness and stealthiness while remaining simple and cost-effective. These features enable its application across a wide range of ML models and scenarios with minimal modifications.


\section{Experiments}
In this section, we conduct extensive experiments to shed light on \textbf{1)} The effectiveness and stealthiness of \textsc{XSub} in attacking image classifiers, \textbf{2)} The  impact of hyper-parameters $\alpha$, $\beta$, and $K$ on the attack's effectiveness, and \textbf{3)} The robustness of our attack against defenses. 



\subsection{Baselines and ML Explainer}
To evaluate our attack \textsc{XSub} and compare it with \textit{1)} \textit{EG-Booster} \cite{amich2022eg}, which is one of the state-of-the adversarial attack against image classifiers,  \textit{2)} \textit{Clean} model, which refers to the original model without any attacks or defenses, and \textit{3)} the defense method \textit{Beatrix} \cite{ma2022beatrix} to assess how well our attack performs against a defense. 

In our experiments, we focus on image datasets and neural network models. Therefore, we choose SHAP \cite{lundberg2020local,lundberg2017unified}, which has been shown to be effective in explaining deep neural networks.  Specifically, we employ the SHAP DeepExplainer tool, which is tailored by SHAP authors for deep learning models in image classification tasks. In addition, SHAP has no access to the target model, which makes  \textsc{XSub} well-suited for the  threat models discussed.

\begin{figure*}[t]
 \centering
\subfigure[CIFAR-10 dataset]{\includegraphics[scale=0.42]{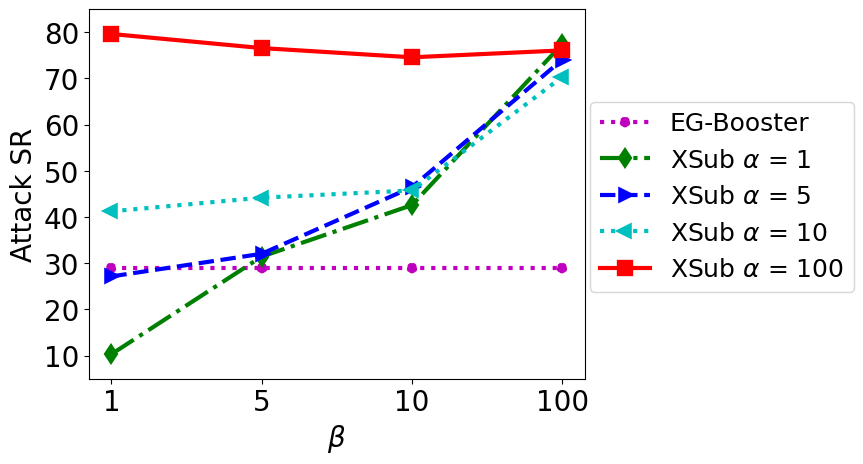}}\hfill
\subfigure[Imagenette dataset]{\includegraphics[scale=0.42]{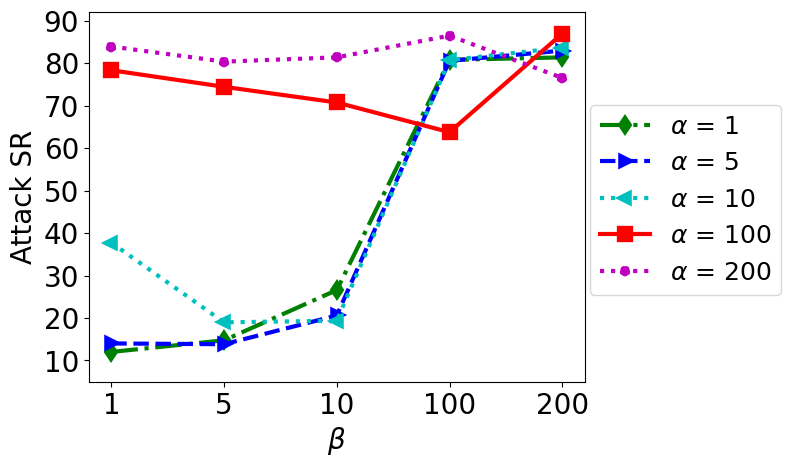}}\hfill
\caption{Attack SR  at different values of $\alpha$ and $\beta$ ($K=1$).} 
\label{fig:AttackSR}
\end{figure*}
\setlength{\textfloatsep}{5pt}

\subsection{Datasets and Model Configurations}

We evaluated our attack on benchmark image classifier datasets, including  CIFAR-10  \cite{krizhevsky2009learning} and  Imagenette \cite{Howard_Imagenette_2019}. The CIFAR-10 dataset has $50,000$ training samples, each having $32 \times 32\times 3$ pixels.  The Imagenette dataset has $9,469$ images in its training set, with each image being $128 \times 128\times 3$ pixels. There are $10$  classes in each dataset.
To pre-process the data, we scaled all pixel values to the range $[0,1]$. Each image was then normalized using Z-score normalization. For the CIFAR-10 dataset, we used means of $\{0.4914, 0.4822, 0.4465\}$ and  standard deviations of $\{ 0.2023, 0.1994, 0.2010 \}$ for the  red, green, and blue color channels, respectively. For the Imagenette dataset, the mean values are $\{0.485, 0.456, 0.406\}$, and the standard deviation values are $\{0.229, 0.224, 0.225\}$.

To evaluate the attacks and avoid counting the Clean model's misclassification of clean data as the attack's success, we only consider samples that were correctly classified by the Clean model. Therefore, the  testing sets used for the CIFAR-10 and Imagenette datasets are $9,806$ and  $3,925$, respectively. In addition, the golden image for each class was  selected among these correctly classified images. The golden image selection for each class was performed offline to reduce repeated queries for a single data sample. 

For the CIFAR-10 dataset, we used a convolutional neural network  with four convolutional layers, three fully connected layer together with 0.25-dropout and max-pooling layers \cite{jameschengpeng2019}. SGD optimizer was implemented together with a learning rate of 0.01. For the Imagenette dataset, we used a XResNet50-based model \cite{fastai_xresnet, imagenette_with_pytorch}, with a Ranger optimizer and a learning rate of 0.008. 
 All experiments are run $10$ times, and the results are reported as the average.

\subsection{Evaluation metrics}
We evaluate our attack with image classification tasks using \textit{1)} Qualitative evaluation, by visualizing images before and after being attacked or under different attack scenarios, and  \textit{2)} Quantitative metrics, including model accuracy and attack success rate, as follows:
\begin{align}
    \text{Accuracy} &=  \frac{ \sum_{i=1}^{N_{test} }\mathbb{I} \Big( f(x_i) = y_i \Big)}{N_{test}} \\
    \text{Attack SR} & = \frac{ \sum_{i=1}^{N_{test} }\mathbb{I} \Big( f(x'_i) \neq y_i \Big)}{N_{test}}
\end{align}
where $N_{test} $ is the total number of testing data samples, and  $\mathbb{I}(\cdot)$ is the indicator function in which  $\mathbb{I}(x)=1$ if $x$ is True and $\mathbb{I}(x)=0$ if $x$ is False. Here, $y_i$ is the ground-truth label of $x_i$ and $x'_i$ is the poisoned sample of $x_i$. Intuitively, the higher Attack SR indicates a more effective attack. In addition, in a  backdoor attack setting, a smaller gap between the Accuracy of the Clean model and the attacked model signifies a more effective attack.

To evaluate our attack and compare with other baselines, we tested a wide rage of hyper-parameters, including $\alpha \in \{1, 5, 10, 100,200\}$, $\beta \in \{1, 5, 10, 100,200\}$, and $K \in \{1,5,30,60,90,120 \}$.

\subsection{Evaluation Results and Discussions}

In the CIFAR-10  and  Imagenette datasets, the Accuracy of the Clean model is  $82\%$ and $88\%$, respectively.  These values are considered the upper bounds for model utility, i.e., Accuracy,  on each dataset.

\textit{\textbf{Adversarial Attack.}} Figure \ref{fig:AttackSR} illustrates the Attack SR of \textsc{XSub} as a function of $\alpha$, $\beta$ with $K=1$ with the CIFAR-10 and Imagenette datasets. In the CIFAR-10 dataset (Figure \ref{fig:AttackSR}a), as  $\alpha$ and  $\beta$ increase, the Attack SR significantly increases, especially when $\alpha$ is small (i.e., $\alpha \in [1,5]$). 
For instance, with $\alpha \in [1,5]$, when $\beta$ increases from $1$ to $100$, the Attack SR increases by $46.92\%-67.14\%$. The gap is smaller when $\alpha$ is larger. When both $\alpha$ and $\beta$ are high, i.e., $\alpha = \beta = 100$, the Attack SR is high, ranging from $74.06\%-79.62\%$. When $\alpha$ or $\beta$ are larger than $5$, \textsc{XSub} outperforms EG-Booster, which has an Attack SR of only  $28.87\%$ \cite{amich2022eg}. 
Note that \textsc{XSub} not only achieves a higher Attack SR but also operates as a black-box model, eliminating the need for multiple requests to the prediction model as required by EG-Booster. The Imagenette dataset (Figure \ref{fig:AttackSR}b) follows the similar trend as in the CIFAR-10 dataset in which Attack SR generally increases when $\alpha$ and $\beta$ increase. However, the results for this dataset exhibit  more fluctuation. This may be due to the Imagenette's significant higher resolution compared with the CIFAR-10 dataset, with $128\times 128$ pixels versus $32\times 32$ pixels. As a result, perturbing a single pixel in the larger image may introduce greater variability, leading to more fluctuating outcomes.

Intuitively, given the fixed value of $K$, as $\alpha$ and $\beta$ increase, the important features of the ground-truth label in the original sample, indicated by the model explanations, are replaced by important features of other labels. 
This substitution causes the model's focus to deviate from the correct class, diminishing its ability to recognize the ground-truth label accurately. 
With high values of $\alpha$ and $\beta$, this replacement effect is exacerbated. The model begins to emphasize irrelevant or misleading features, which significantly increases the probability of misclassification. 
 Meanwhile, the Accuracy exhibits a slight decrease across all values of $\alpha$ and $\beta$, compared with that of the Clean model.  The model becomes more susceptible to errors in classification, reflecting a trade-off between the attack effectiveness and  model utility.

\begin{figure}[t]
 \centering
\subfigure{\includegraphics[scale=0.25]{viz/pert_img/overview.png}}\hfill
\subfigure{\includegraphics[scale=0.25]{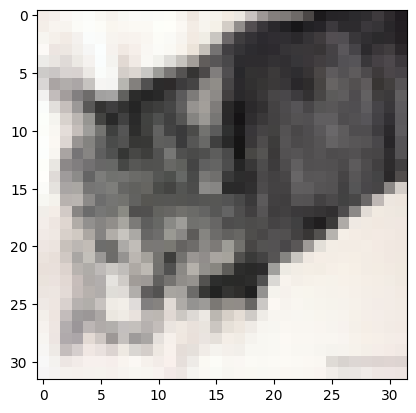}}\hfill
\subfigure{\includegraphics[scale=0.25]{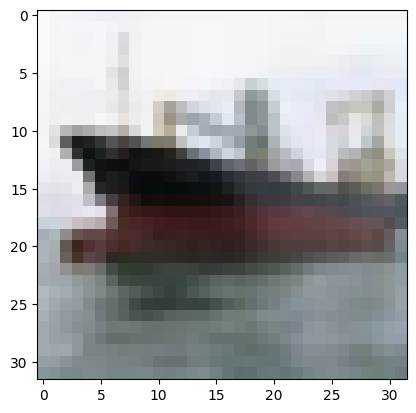}}\hfill
\subfigure{\includegraphics[scale=0.25]{viz/pert_img/overview_k1.png}}\hfill
\subfigure{\includegraphics[scale=0.25]{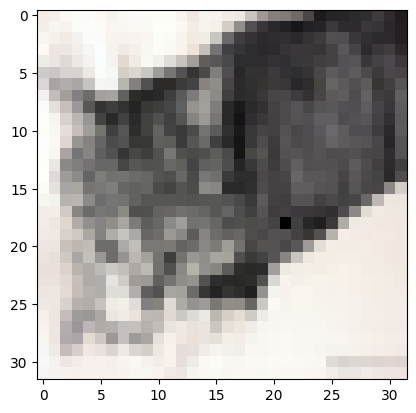}}\hfill
\subfigure{\includegraphics[scale=0.25]{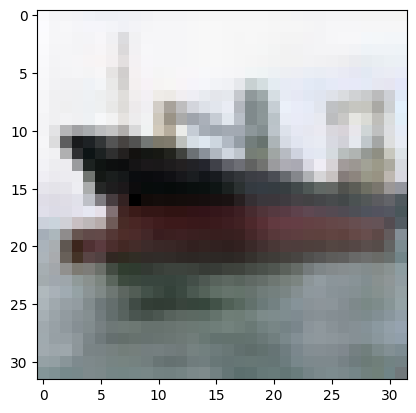}}\hfill
\subfigure{\includegraphics[scale=0.25]{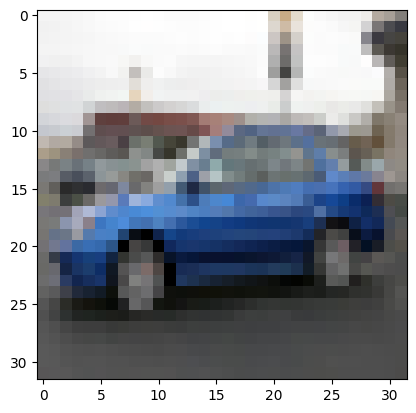}}\hfill
\subfigure{\includegraphics[scale=0.25]{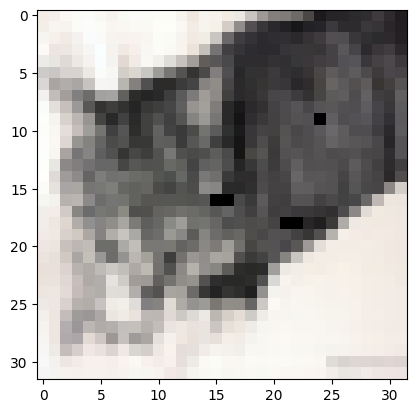}}\hfill
\subfigure{\includegraphics[scale=0.25]{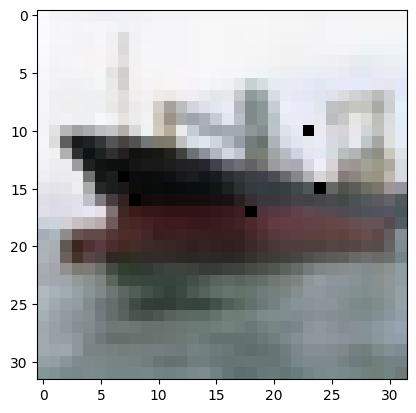}}\hfill
\subfigure{\includegraphics[scale=0.25]{viz/pert_img/overview_k30.png}}\hfill
\subfigure{\includegraphics[scale=0.25]{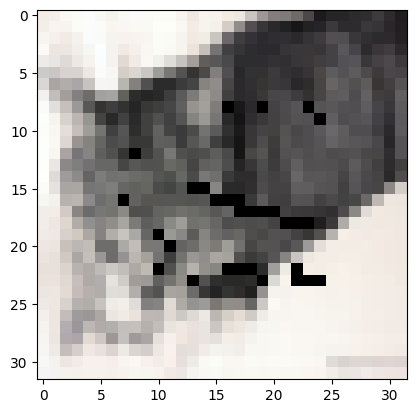}}\hfill
\subfigure{\includegraphics[scale=0.25]{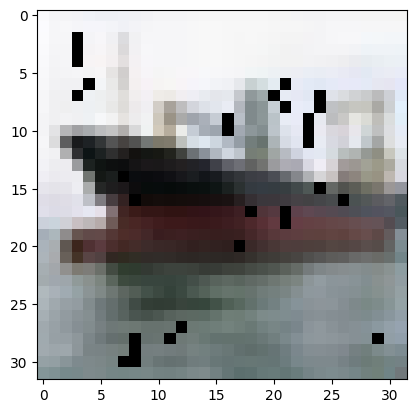}}\hfill
\caption{Original images (top row) and their poisoned versions after being attacked by  \textsc{XSub} in the CIFAR-10 dataset with different values of $K$ ($K=1$ in the second row, $K=5$ in the third row, and $K=30$ in the last row).} 
\label{fig:visualizationK}
\end{figure}

\begin{figure}[t]
 \centering
\subfigure{\includegraphics[scale=0.25]{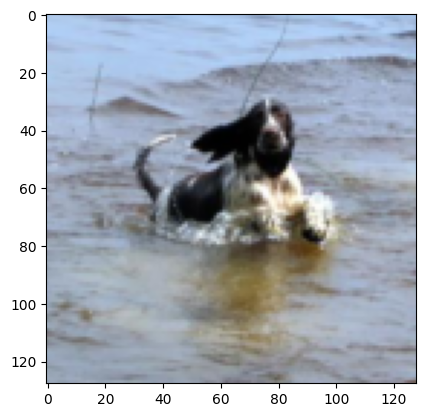}}\hfill
\subfigure{\includegraphics[scale=0.25]{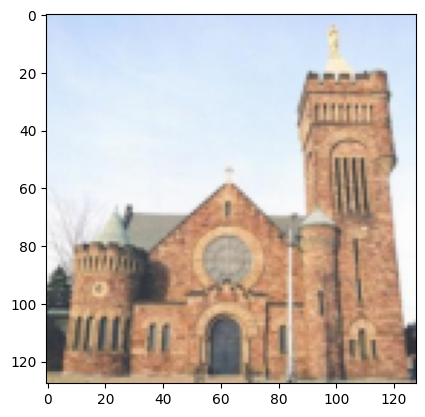}}\hfill
\subfigure{\includegraphics[scale=0.25]{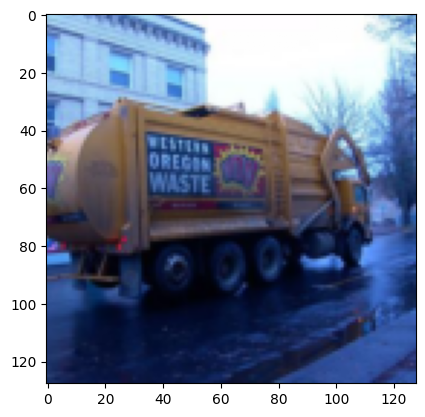}}\hfill
\subfigure{\includegraphics[scale=0.25]{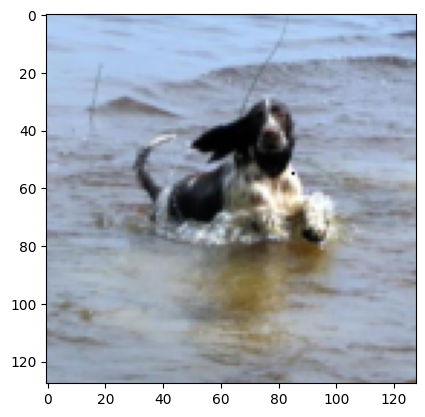}}\hfill
\subfigure{\includegraphics[scale=0.25]{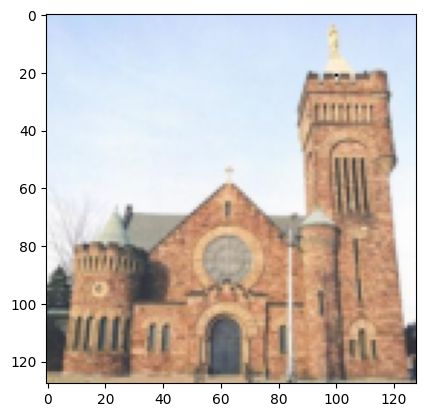}}\hfill
\subfigure{\includegraphics[scale=0.25]{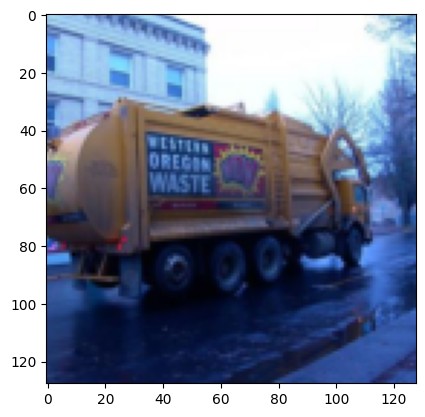}}\hfill
\subfigure{\includegraphics[scale=0.25]{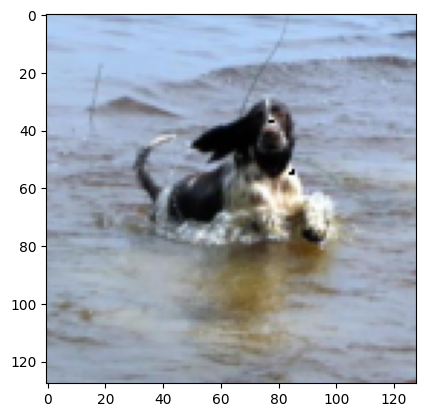}}\hfill
\subfigure{\includegraphics[scale=0.25]{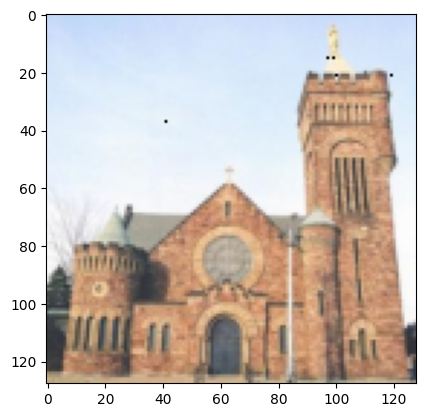}}\hfill
\subfigure{\includegraphics[scale=0.25]{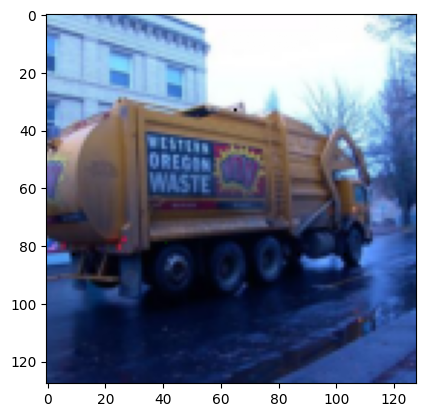}}\hfill
\subfigure{\includegraphics[scale=0.25]{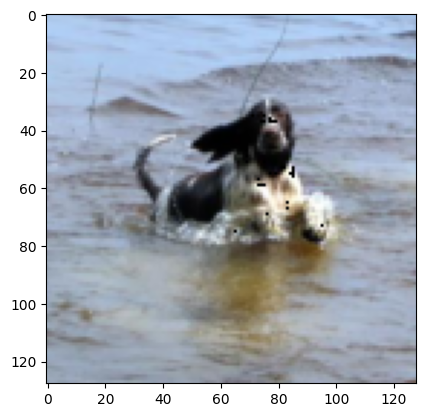}}\hfill
\subfigure{\includegraphics[scale=0.25]{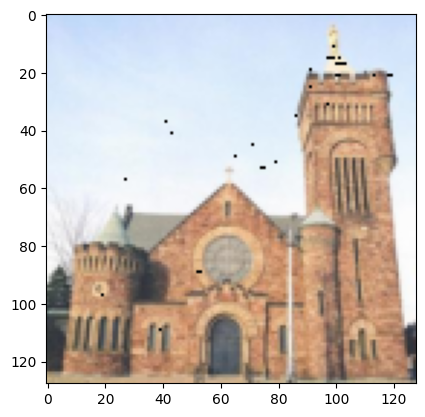}}\hfill
\subfigure{\includegraphics[scale=0.25]{viz/pert_img/truck_K5.png}}\hfill
\caption{Original images (top row) and their poisoned versions after being attacked by  \textsc{XSub} in the Imagenette dataset with different values of $K$ ($K=1$ in the second row, $K=5$ in the third row, and $K=30$ in the last row).} 
\label{fig:visualizationK-imagenette}
\end{figure}

For qualitative evaluation, we visualize images before and after being attacked by our attack \textsc{XSub}. As shown in Figures  \ref{fig:visualizationK} and \ref{fig:visualizationK-imagenette}, when $K=1$, only a single feature is affected by our attack, resulting in minimal differences between the images before and after the attack, which are less noticeable to the human eye. Even with different values of $\alpha$ and $\beta$, these differences remain subtle. When $K$ increases,  more features are replaced, making the differences  more obvious. For instance, in the CIFAR-10 dataset, starting from $K=30$, the malicious patterns become.   However, with the Imagenette dataset, due to its higher pixel resolution, these differences remain less noticeable even at $K=30$.  It is noteworthy that with $K=1$, we already achieve a high Attack SR (greater than $70\%$ in the CIFAR-10 dataset or greater than $80\%$ in the Imagenette dataset).

\begin{figure}[t]
      \centering
       \includegraphics[scale=0.42]{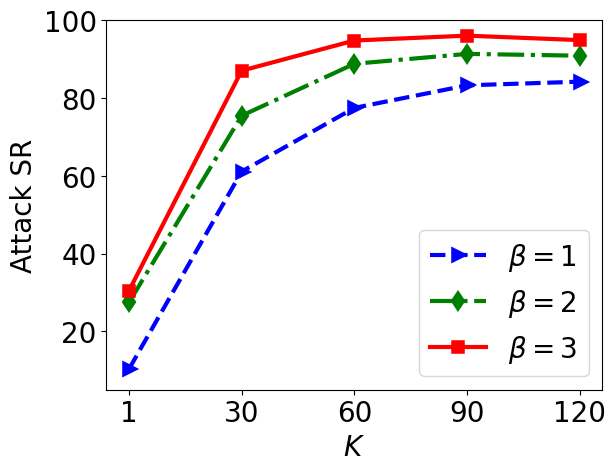} 
      \caption{Attack SR at different values of $K$ ($\alpha =1$) in the CIFAR-10 dataset.}
      \label{fig:AttackSR-k}
\end{figure}

\textit{\textbf{Impacts of $\alpha$, $\beta$, and $K$ on Attack Effectiveness and Stealthiness.}} 
Figures \ref{fig:AttackSR} and \ref{fig:AttackSR-k} illustrate that $\alpha$, $\beta$, and $K$ have substantial impacts on Attack SR, with $K$ showing particularly strong effects.  When $\alpha$, $\beta$, and $K$ increase, the Attack SR rises notably. 
For instance, in the CIFAR-10 dataset, with $K=1$, increasing $\alpha$ from $1$ to $100$ results in a substantial uplift in  Attack SR, from $10.30\%$ to $79.62\%$. For values of  $\alpha \in [1,10]$, increasing $\beta$ also greatly increases  Attack SR with $67.14\%$ uplift. However, when  $\alpha$ is sufficiently high, i.e., $\alpha = 100$,  the impact of $\beta$ is subtle, with  Attack SR fluctuating between $74.57\%-79.62
\%$ as $\beta \in [1,100]$. 
When $K>1$, even with a small value of $\alpha$ and $\beta$, i.e., $\alpha = 1$ and $\beta \in \{ 1,2,3\}$, Attack SR significantly improves compared with that of $K=1$. For instances, at $K=90$, $\alpha=1$, and $\beta=3$, Attack SR can reach $96.01\%$. This is because increasing $K$ results in more important features of the ground-truth label being replaced by features from other labels, which increases the likelihood of misclassification. However, we also observe that when $K$ is sufficiently high, i.e., $120$,  Attack SR tends to decrease. This reduction is because, with high values of $K$, the replacement for high order features, i.e., features $91^{th}$ to $120^{th}$, which are less important, becomes more random. Such random replacement can lead to confusion rather than misclassification, which results in a slight drop in Attack SR. 
We observe similar trends in the Imagenette dataset.


However, visualizing the images before and after the attack (Figures \ref{fig:visualizationK} and \ref{fig:visualizationK-imagenette}) clearly demonstrates the trade-off between attack effectiveness and stealthiness, especially when $K$ is large. With a large value of $K$, many pixels are affected, making the changes visible to the human eye and easier to detect. Therefore, for further experiments, we focus on $K=1$, which offers the best balance between the effectiveness and stealthiness of \textsc{XSub}.

\begin{table}[t] 
\centering 
\caption{\textsc{XSub} against the Beatrix defense on the CIFAR-10 dataset ($K=1$).} 
\begin{tabular}{|c|c|c|c|c|c|}
 \hline
$\alpha$  &  $\beta$  &   Detection rate \\ 
 \hline
  \hline
\multirow{2}{*}{1} & 10   & 8.68\% \\
    & 100 &   49.26\% \\
                    \hline
\multirow{2}{*}{5} & 10  & 11.68\% \\
    & 100 &   52.02\% \\
                    \hline
\multirow{2}{*}{10} & 10 &    12.40\% \\
    & 100 &    55.56\% \\           
 \hline
\end{tabular}
\label{tab:defense-cifar10}  
\end{table}
\setlength{\textfloatsep}{5pt}

\textit{\textbf{Extension to Backdoor Attack.}} 
To further extend our attack to a backdoor setting where the attacker can have (limited) access the training data of the ML model, we poison the training samples in a manner akin to the adversarial attack setting, as outlined in Algorithm \ref{alg: LDPstealIP}. These poisoned samples are then used to train the ML model, similar to ML-as-a-Service scenarios that utilize crowd-sourced data. This approach allows attackers to manipulate the training data, thereby compromising the integrity of the model.


For this experiment, we randomly select $10\%$ of the training data, e.g., $5,000$ images in the CIFAR-10 dataset or $1,000$ images in the Imagenette dataset and apply \textsc{XSub} to poison them. For each image chosen for poisoning, we select a random label different from its ground-truth label as a target for perturbation. Using the setting that achieved the highest Attack SR in the adversarial attack setting, i.e., $K=1$, $\alpha=100$, and $\beta=1$, we achieve an  Attack SR of $75.54\%$ in the CIFAR-10 dataset and $75.31\%$ in the Imagenette dataset, reflecting slight decreases of $4.08\%$ and $5.01\%$ from the adversarial settings. The subsequent backdoored model exhibits an Accuracy of $74.05\%$ in the CIFAR-10 dataset and $82.04\%$ in the Imagenette dataset. 
Basically, we perturb the samples from a class to another  random untargeted class, which can influence the decision boundary of all classes (not only a certain class as in targeted backdoor attacks), causing a moderate drop in accuracy. In addition, during backdoor training, the decision boundary  shifts to accommodate both clean and poisoned data, resulting in a slight decrease in Attack SR compared with the adversarial setting, which does not affect the ML model. 

%

 \textit{\textbf{\textsc{XSub} against Defenses.}} 
 To further evaluate the robustness of our attack, we examine the effectiveness of \textsc{XSub} against the Beatrix defense  \cite{ma2022beatrix}. We adopt the detection threshold specified in the paper, set at  $99\%$.  This threshold indicates that $99\%$  of the data samples in the benign training dataset have median absolute deviations smaller than the chosen detection threshold. 
The detection rate is calculated as the percentage of poisoned samples that are correctly detected as poisoned samples.
 A lower detection rate indicates greater robustness of our attack.  As shown in Table \ref{tab:defense-cifar10}, our attack achieves a sufficiently low detection rate with small values of $\beta$, i.e., $8\%$ detection rate at $\beta=10$. When $\alpha$ increases, the detection rate slightly rises, with an uplift of  $3.72\%$.

\setlength{\textfloatsep}{5pt}

\setlength{\textfloatsep}{5pt}

 \section{Conclusion}

This paper introduced \textsc{XSub}, a novel explanation-driven adversarial attack against  black-box classifiers by leveraging feature substitution.  
The key concept behind \textsc{XSub} involves perturbing the most important features that are identified by an explainer. Specifically, an attacker would strategically substitute important features in the original sample with corresponding features from the golden sample, which is a sample from a different class that contains the most influential feature for that class. By doing that, our attack significantly increases the probability of the model incorrectly classifying the perturbed samples.
This method allows for precise control over the trade-off between attack effectiveness and stealthiness, making \textsc{XSub} both a potent and adaptable tool for various attack scenarios. Additionally, \textsc{XSub} maintains a constant query complexity. Its cost-effectiveness and ease of adaptation to backdoor attacks further highlight its potential impact. Our experiments show that \textsc{XSub} outperforms existing attacks and is robust against defense mechanisms. Our research reinforces and highlights a security trade-off of XAI in that it promotes transparency while simultaneously revealing more information to adversaries, making black-box models more vulnerable to attacks. This calls for future research in addressing this security trade-off.

\section*{Acknowledgment}

This work was authored in part by the National Renewable Energy Laboratory, operated by Alliance for Sustainable Energy, LLC, for the U.S. Department of Energy (DOE) under Contract No. DE-AC36-08GO28308. This work was supported by the State University of New York at Albany (SUNY Albany) under IFR 940008-20, by the U.S. Department of Energy Office of Cybersecurity, Energy Security, and Emergency Response (CESER), and by the Laboratory Directed Research and Development (LDRD) Program at NREL. The views expressed in the article do not necessarily represent the views of the DOE or the U.S. Government. The U.S. Government retains and the publisher, by accepting the article for publication, acknowledges that the U.S. Government retains a nonexclusive, paid-up, irrevocable, worldwide license to publish or reproduce the published form of this work, or allow others to do so, for U.S. Government purposes.

\bibliographystyle{IEEEtran}
\bibliography{{IEEEfull}}

\end{document}